\documentclass[letterpaper]{article} % DO NOT CHANGE THIS
\usepackage[hyphens]{url}  % DO NOT CHANGE THIS
\usepackage{graphicx} % DO NOT CHANGE THIS
\usepackage{natbib}  % DO NOT CHANGE THIS AND DO NOT ADD ANY OPTIONS TO IT
\usepackage{caption} % DO NOT CHANGE THIS AND DO NOT ADD ANY OPTIONS TO IT
\usepackage{amsmath}
\usepackage{amssymb}
\usepackage{booktabs}
\usepackage{multirow}
\usepackage{bm}

\usepackage{eovsam2027}  % 移除 [submission] 选项以显示作者
\nocopyright % 添加这一行来彻底移除左下角的版权声明

\usepackage{xcolor}

\title{EOVSAM: Efficient Open-Vocabulary Segmentation with SAM 3 in One Pass}

\author{
    Haomin Peng\textsuperscript{1}, 
    Yongkang Li\textsuperscript{1}, 
    Zhaoxiang Liu\textsuperscript{2,3}, 
    Xiaojie Jin\textsuperscript{4},
    Shiguo Lian\textsuperscript{2,3}, 
    Yunchao Wei\textsuperscript{4}, 
    Xinggang Wang\textsuperscript{1}\thanks{Corresponding author.} \\
}

\affiliations{
    \textsuperscript{1}Huazhong University of Science and Technology \quad
    \textsuperscript{2}Data Science \& Artificial Intelligence Research Institute,  China Unicom \quad
    \textsuperscript{3}Unicom Data Intelligence, China Unicom \quad
    \textsuperscript{4}Beijing Jiaotong University 
}

\begin{document}

\maketitle

\begin{abstract}
Open-vocabulary segmentation identifies and segments objects from arbitrary textual descriptions. SAM 3 supports noun-phrase-guided segmentation and achieves competitive open-vocabulary performance through exhaustive vocabulary traversal, yet suffers from prohibitive computational overhead as target categories scale.
In this paper, we propose an Efficient Open-Vocabulary segmentation framework with SAM 3 (EOVSAM), which adapts SAM 3 for single-pass prediction. EOVSAM removes prompt conditioning to turn SAM 3 into an efficient mask generator and introduces a new Attentional Aggregation strategy to optimize open-vocabulary classification end-to-end. This formulation avoids the multi-stage pipelines and post-processing heuristics commonly used by existing methods, while mitigating the closed-set collapse that can arise when classification is optimized directly. EOVSAM consistently improves segmentation accuracy over vanilla SAM 3 on all evaluated datasets and accelerates inference by up to 338$\times$. Furthermore, EOVSAM maintains high accuracy at lower resolutions while achieving even more remarkable inference speeds. Experiments on standard semantic and panoptic segmentation benchmarks show that EOVSAM combines competitive or state-of-the-art accuracy with a substantial speed advantage over existing open-vocabulary segmentation models. Code and models are available at 
\textcolor{blue}{\texttt{https://github.com/hustvl/EOVSAM}}.
\end{abstract}

%     \link{Code}{https://eovsam.org/example/code}
%     \link{Datasets}{https://eovsam.org/example/datasets}
%     \link{Extended version}{https://eovsam.org/example/extended-version}

\section{Introduction}

\begin{figure}[!t]
    \centering
    \includegraphics[width=1.0\linewidth]{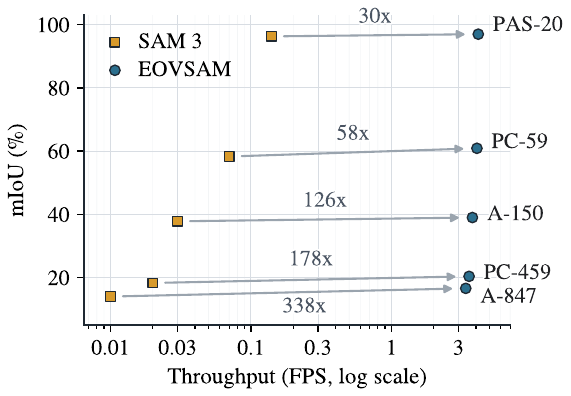}
    \caption{
Inference speed and segmentation performance on datasets with different vocabulary sizes.
EOVSAM operates at a resolution scale of 1152, compared to 1008 for SAM 3.
    }
    \label{fig:speed_comparison}
\end{figure}

Conventional image segmentation methods~\cite{deeplabv3plus2018,he2017mask,mask2former} generally assume a closed set of categories defined by their training data. Because dense pixel-level annotations are expensive to obtain, most segmentation datasets cover a limited semantic range, which constrains the use of these models in real-world scenes. Open-vocabulary segmentation (OVS) addresses this limitation by using text to specify the objects to segment, rather than relying solely on category representations learned from a fixed training vocabulary. It can therefore extend segmentation to categories not observed during training.

Recent work has explored how the Segment Anything Model (SAM) family~\cite{sam,sam2,SAM_3} can support open-vocabulary segmentation. Existing approaches typically either integrate SAM features or use SAM through prompts. Feature-integration methods~\cite{ebseg,frozenseg} fuse features from a SAM image encoder into the mask generator of an existing open-vocabulary pipeline, but the resulting gains are often modest. Prompt-driven methods~\cite{sam-mi,openworldsam} instead generate pseudo-prompts and process them with SAM. Because each prompt is handled independently, this approach adds considerable computation. SAM 3~\cite{SAM_3} natively supports noun-phrase-guided segmentation and returns a fixed number of candidate masks with confidence scores for each phrase. To perform open-vocabulary segmentation, it processes every category in the target vocabulary separately and merges the predictions through post-processing. This procedure is accurate, but its inference cost increases sharply with vocabulary size, as shown in Figure~\ref{fig:speed_comparison}.

We propose EOVSAM, an efficient open-vocabulary segmentation framework built directly on SAM 3. To remove the cost of iterative inference, we adapt SAM 3 into a prompt-free mask generator. The agglomerative vision encoder C-RADIOv4~\cite{c_radio_v4} extracts both SAM 3 and SigLIP 2~\cite{SigLIP_2} features. After the fusion encoder refines the SAM 3 features, the detector decoder produces mask embeddings together with classification-oriented attention maps. These maps aggregate the SigLIP 2 features into object embeddings for open-vocabulary recognition.

This design offers several key practical advantages. First, the prompt-free SAM 3 mask generator produces high-quality masks at a lower computational cost than prior approaches based on Mask2Former~\cite{mask2former}. Second, Attentional Aggregation jointly optimizes mask localization and open-vocabulary recognition in a single pass. In contrast to methods that classify masks only after generating them, this end-to-end formulation allows the decoder to learn from the recognition objective. Furthermore, EOVSAM maintains highly competitive accuracy even at substantially reduced input resolutions, yielding further inference acceleration of over 2$\times$ compared to its default setting. Ultimately, by coupling our efficient single-pass architecture with this zero-shot scaling capability, EOVSAM provides a highly flexible trade-off between latency and performance, making the overall framework uniquely suited for latency-sensitive and resource-constrained deployment.

We evaluate EOVSAM on several open-vocabulary semantic and panoptic segmentation benchmarks~\cite{ade20k,pascal_context,Pascal-VOC}. EOVSAM substantially reduces inference latency and outperforms vanilla SAM 3 on every semantic segmentation benchmark considered. Among models trained for open-vocabulary panoptic segmentation, it achieves state-of-the-art semantic segmentation results on A-150~\cite{ade20k}, A-847, PC-459~\cite{pascal_context}, and PAS-20~\cite{Pascal-VOC}, while remaining competitive with methods specialized for semantic segmentation. It also sets a new state of the art for open-vocabulary panoptic segmentation on ADE20K.
Our main contributions are summarized as follows:
\begin{itemize}
\item We present a novel, single-pass paradigm for adapting SAM 3 to open-vocabulary segmentation. By transforming the architecture into a prompt-free mask generator, the proposed framework effectively leverages SAM 3's powerful localization capabilities while entirely eliminating the prohibitive multi-pass computational bottleneck.

\item We introduce Attentional Aggregation, an end-to-end joint optimization mechanism that elegantly resolves open-vocabulary classification challenges. By maintaining a continuous gradient flow and unifying mask localization with recognition, it achieves both high efficiency and accuracy.
\end{itemize}

\section{Related Work}

\subsection{Open-Vocabulary Segmentation}

Open-vocabulary segmentation uses arbitrary text descriptions to segment objects, including categories not seen during training. Existing methods can be grouped by what they classify. Pixel-based methods assign a category to each pixel directly, whereas mask-based methods classify a set of mask proposals.
FC-CLIP~\cite{fcclip} and MAFT-Plus~\cite{maftplus} avoid running CLIP separately on every masked region by adopting a convolutional CLIP visual backbone and pooling its image features within each mask.
DeOP~\cite{deop} uses heatmap-based pooling, while Mask-Adapter~\cite{maskadapter} replaces mask pooling with a semantic activation map generator that captures more contextual information. Mask-Adapter improves accuracy but also adds substantial inference latency.
Pixel-based methods~\cite{catseg,sed,fgaseg,xagent,sam-mi,escnet} cannot separate multiple instances of the same category and therefore do not directly support panoptic segmentation.

The Segment Anything Model (SAM)~\cite{sam,sam2,SAM_3} family provides strong general-purpose segmentation models.
Several studies have adapted SAM to open-vocabulary segmentation. Among mask-based methods, EBSeg~\cite{ebseg} uses a linear layer to fuse CLIP and SAM image features. FrozenSeg~\cite{frozenseg} extends FC-CLIP~\cite{fcclip} by injecting SAM features into both the decoder queries and the CLIP visual features, and ensembles zero-shot SAM masks at inference time to improve mask proposals. Among pixel-based methods, SAM-MI~\cite{sam-mi} pools SAM masks into region-level embeddings and incorporates them into cost aggregation.
% ESC-Net~\cite{escnet} further introduces correlation-based pseudo-prompts derived from CLIP image-text features and injects them into SAM’s prompt encoder, enabling pretrained SAM transformer blocks to enhance image features.
SAM 3~\cite{SAM_3} natively supports text-driven concept segmentation through vision-language perception encoder pretraining and detector training on noun-phrase--mask pairs. However, it cannot process all category prompts jointly, making open-vocabulary inference computationally expensive. We instead adapt SAM 3 for efficient single-pass open-vocabulary segmentation.

\begin{figure*}[t!]
    \centering
    \includegraphics[width=1.0\linewidth]{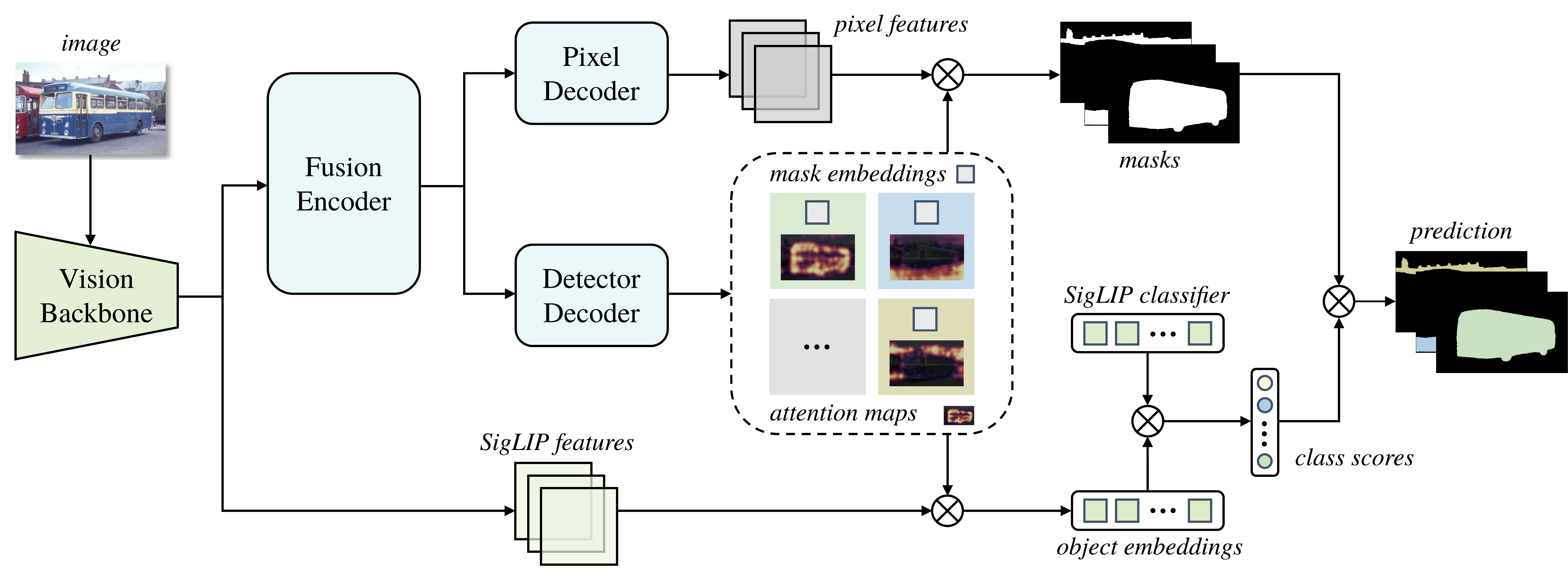}
    \caption{Adapting SAM 3 for open-vocabulary segmentation. EOVSAM replaces the SAM 3 image encoder with an agglomerative vision backbone that extracts SAM and SigLIP features in parallel. The detector decoder produces mask embeddings and corresponding attention maps. These maps aggregate the SigLIP features into object embeddings aligned with the predicted masks, which are classified by their cosine similarity to the SigLIP text classifier. This single-pass design removes SAM 3's repeated inference over the target vocabulary.}
    \label{fig:architecture}
\end{figure*}

\subsection{Vision Foundation Models}
Vision-language models, such as CLIP~\cite{clip} and ALIGN~\cite{align}, demonstrate that pre-trained dual-encoder models can learn cross-modal aligned representations and show strong performance on zero-shot downstream tasks. SigLIP~\cite{siglip} proposes a simple pairwise sigmoid loss for language-image pre-training. Building upon this, SigLIP 2~\cite{SigLIP_2} introduces captioning-based pre-training, self-supervised losses (e.g., self-distillation and masked prediction), and online data curation, leading to significant improvements on localization and dense prediction tasks.
The Segment Anything Model (SAM)~\cite{sam} serves as a segmentation-oriented vision foundation model. Trained on the massive SA-1B dataset comprising one billion masks across 11 million images, SAM achieves high-precision segmentation guided by geometric prompts, such as points and bounding boxes. Building upon this, SAM 3~\cite{SAM_3} further supports short text phrases as prompts. Distinct from typical vision-language models, SAM 3 utilizes textual prompts merely as conditional inputs fed into the decoder via cross-attention. Consequently, the visual and textual features in SAM 3 do not explicitly reside within an aligned joint feature space.
AM-RADIO~\cite{am_radio} introduces the concept of agglomerative foundation models, a paradigm that creates a unified foundation model by distilling feature representations from multiple heterogeneous architectures. While the original formulation utilizes DFN CLIP, DINOv2, and SAM as its core teacher set, C-RADIOv4~\cite{c_radio_v4} builds upon the AM-RADIO and RADIOv2.5~\cite{radio_v2_5} frameworks by updating the teacher set to SigLIP 2, DINOv3~\cite{dinov3}, and SAM 3. 
% Furthermore, C-RADIOv4 possesses the excellent characteristics of supporting variable resolutions and serving as a direct replacement for the SAM 3 vision encoder.

\section{Approach}

\subsection{Problem Definition}
Open-vocabulary segmentation aims to segment images into semantic categories defined by arbitrary textual descriptions, rather than being restricted to a fixed set of predefined classes. Typically, a model is trained on a dataset containing a set of seen categories, $\mathcal{C}_\text{seen}$, and is expected to generalize to both $\mathcal{C}_\text{seen}$ and an open set of unseen categories, $\mathcal{C}_\text{unseen}$, during inference on novel datasets.

The recently proposed SAM 3 leverages the large-scale SA-Co dataset to excel at Promptable Concept Segmentation (PCS), a task involving the segmentation of all instances of a visual concept specified by a short text phrase. However, it is inherently designed for single-concept prompts. Consequently, SAM 3 can only generate masks for one category per forward pass. Adapting this paradigm to open-vocabulary semantic or panoptic segmentation tasks necessitates multi-pass inference and relies on heavy post-processing heuristics. While this costly approach yields acceptable semantic accuracy, its failure to maintain instance-level exclusivity severely degrades panoptic segmentation performance.

\subsection{Adapting SAM 3 for Single-Pass Full-Image Segmentation}

To better understand our architectural modifications, we first briefly revisit the standard segmentation pipeline of vanilla SAM 3. Given an input image $I$ and a specific text prompt describing a target concept, SAM 3 extracts unconditioned image features via an image encoder and text features via a text encoder. A fusion encoder then conditions the image features by cross-attending to the text features. In the detector decoder, the learnable queries compute cross-attention with the conditioned image features and text features, respectively, and subsequently perform a dot product with the high-resolution feature maps produced by the pixel decoder to yield the predicted masks. Thus, open-vocabulary segmentation over a vocabulary of size $K$ forces the model to perform $K$ independent forward passes. This repeated computation severely limits its efficiency.

Despite this multi-pass bottleneck, we observe that the detector decoder follows the general DETR paradigm, which inherently provides the structural foundation for single-pass segmentation. Recognizing this, our framework focuses on unlocking this latent potential through a minimalist architectural adaptation. The designs of the fusion encoder, pixel decoder, and detector decoder largely preserve the original structural integrity of SAM 3, enabling the seamless integration of its pre-trained weights. Specifically, we remove the text cross-attention layers from the fusion encoder and detector decoder, eliminating the model's reliance on prompts for prediction while minimizing computational redundancy.

Our framework must simultaneously address two key challenges: precise mask localization and robust semantic recognition. As illustrated in Figure~\ref{fig:architecture}, to efficiently extract representations tailored for both tasks without the computational burden of multiple vision encoders, we employ C-RADIOv4~\cite{c_radio_v4} as the vision backbone. Featuring a Vision Transformer equipped with two lightweight adapter heads, C-RADIOv4 extracts parallel features: one tailored for SAM 3, denoted as $\mathcal{F}_\text{sam}$, and another for SigLIP 2, denoted as $\mathcal{F}_\text{siglip}$. For the text modality, a predefined vocabulary of categories is processed by the SigLIP 2 text encoder to generate a SigLIP classifier $\mathcal{T}_\text{siglip}\in\mathbb{R}^{K\times C}$, where $K$ denotes the number of categories and $C$ represents the channel dimension. We also leverage the pre-trained SAM 3 text encoder to encode the entire predefined category vocabulary, which is then appended with a learnable background token to form a SAM classifier $\mathcal{T}_{\text{sam}}\in\mathbb{R}^{(K+1)\times C}$.

In the visual pipeline, the extracted features $\mathcal{F}_\text{sam}$ are first routed into the fusion encoder to obtain enhanced features, denoted as $\mathcal{F}_\text{enh} \in \mathbb{R}^{C \times h \times w}$. $\mathcal{F}_\text{enh}$ are subsequently fed into two separate branches: the detector decoder and the pixel decoder. The detector decoder processes $N$ learnable object queries, denoted as $\mathcal{Q}\in\mathbb{R}^{N\times C}$, along with the enhanced features $\mathcal{F}_\text{enh}$. This decoding process simultaneously yields $N$ mask embeddings $\mathcal{E}_\text{mask}\in\mathbb{R}^{N\times C}$ and their corresponding attention maps $\mathcal{A}\in\mathbb{R}^{N\times h\times w}$, which can be formulated as:
\begin{equation}
(\mathcal{E}_\text{mask}, \mathcal{A}) = \text{DetectorDecoder}(\mathcal{Q}, \mathcal{F}_\text{enh})
\label{eq:decoder_output}
\end{equation}

Simultaneously, the pixel decoder processes $\mathcal{F}_\text{enh}$ to output a high-resolution feature map $\mathcal{F}_\text{seg}\in\mathbb{R}^{C\times H\times W}$. The final semantic masks $\mathcal{M}\in\mathbb{R}^{N\times H\times W}$ are obtained by computing the product between the mask embeddings $\mathcal{E}_\text{mask}$ and the high-resolution features $\mathcal{F}_\text{seg}$:
\begin{equation}
\mathcal{M} = \mathcal{E}_\text{mask} \mathcal{F}_\text{seg}
\label{eq:mask_generation}
\end{equation}

The attention maps $\mathcal{A}$ are used to aggregate $\mathcal{F}_\text{siglip}$ to yield region-specific object embeddings $\mathcal{O}\in\mathbb{R}^{N\times C}$.
We then compute the cosine similarities between these object embeddings $\mathcal{O}$ and the SigLIP classifier $\mathcal{T}_\text{siglip}$ for the final category predictions.
Furthermore, we perform mask pooling over $\mathcal{F}_\text{seg}$ and add the pooled features to $\mathcal{E}_\text{mask}$. The resulting representations are then used to compute cosine similarities with the SAM classifier $\mathcal{T}_{\text{sam}}$ to determine objectness scores and supplementary classification results.
Unlike conventional methods that rely on a segment-then-recognize paradigm, our architecture elegantly unifies mask localization and open-vocabulary recognition into a single-pass joint optimization process.

\begin{table*}[t]
\centering
\small
\setlength{\tabcolsep}{3.8pt}
\begin{tabular}{l | c | ccccc}
% -- 表头部分 --
% 顶部的横线，如果要完全像FC-CLIP那样表头上方没有横线，可以注释掉下面这行
\toprule 
\textbf{Method} & \textbf{Training Dataset} &  A-150 & A-847 & PC-59 & PC-459 & PAS-20 \\
\midrule

% -- COCO-Stuff  --
% R-SC-CLIPself        & COCO-Stuff & 38.4 & 16.6 & 63.6 & -    & -    \\
EBSeg~\cite{ebseg}                & COCO-Stuff & 32.8 & 13.7 & 60.2 & 21.0 & 96.4 \\
MAFT+~\cite{maftplus}                & COCO-Stuff & 36.1 & 15.1 & 59.4 & 21.6 & 96.5 \\
% MROVSeg~\cite{mrovseg}              & COCO-Stuff & 36.9 & 16.4 & \textbf{64.3} & 24.0 & \textbf{97.6} \\
CATSEG~\cite{catseg}               & COCO-Stuff & 37.9 & 16.0 & 63.3 & 23.8 & 97.0 \\
FGAseg~\cite{fgaseg}               & COCO-Stuff & 37.9 & \textbf{16.3} & 63.4 & 23.9 & 97.1 \\
MAFT+ w/ MaskAdapter~\cite{maskadapter} & COCO-Stuff & 38.2 & 16.2 & 60.4 & 22.7 & 95.8 \\
% ERR-Seg~\cite{errseg}              & COCO-Stuff & 38.3 & 16.9 & 60.7 & 23.9 & 96.6 \\
% MaskCLIP++ w/ MAFT+~\cite{zeng2025maskclippp}  & COCO-Stuff & 38.2 & 16.8 & 62.5 & 23.9 & 96.8 \\
X-Agent~\cite{xagent}              & COCO-Stuff & 38.2 & 16.0 & 63.7 & 24.2 & \textbf{97.6} \\
SAM-MI~\cite{sam-mi}               & COCO-Stuff & \textbf{38.6} & \textbf{16.3} & \textbf{63.9} & \textbf{24.4} & 97.2 \\
% ESC-Net~\cite{escnet}              & COCO-Stuff & 41.8 & 18.1 & 65.6 & 27.0 & 98.3 \\
\midrule

% -- COCO Panoptic --
EOV-Seg~\cite{eovseg}              & COCO Panoptic & 32.1 & 12.8 & 56.9 & 16.8 & 94.8 \\
FC-CLIP~\cite{fcclip}              & COCO Panoptic & 34.1 & 14.8 & 58.4 & 18.2 & 95.4 \\
MAFT+$^\dagger$~\cite{maftplus}              & COCO Panoptic & 33.8 & 13.9 & 57.5 & 15.0 & 95.2 \\
FrozenSeg~\cite{frozenseg}            & COCO Panoptic & 34.4 & 14.8 & -    & 19.7 & -    \\
OMTSeg~\cite{omtseg}               & COCO Panoptic & 34.8 & 13.9 & \textbf{61.0} & 17.1 & 95.5 \\
GBA~\cite{gba}                  & COCO Panoptic & 35.9 & 15.1 & 59.6 & 18.5 & 95.8 \\
FC-CLIP w/ MaskAdapter~\cite{maskadapter} & COCO Panoptic & 36.6 & 14.1 & 59.7 & 19.3 & 95.5 \\
% FISA~\cite{fisa}                 & COCO Panoptic & 36.8 & 16.1 & 62.4 & 23.6 & -    \\
% OVSNet~\cite{ovsnet}               & COCO Panoptic & 37.1 & 16.2 & 62.0 & 23.5 & 96.9 \\
OVRCOAT~\cite{ovrcoat}          & COCO Panoptic &  33.7 & 14.0 & 55.4 & 18.1 & 94.1 \\
\midrule

% -- Ours 数据 --
SAM 3$^\dagger$~\cite{SAM_3}                 & SA-Co & 37.8 & 14.0 & 58.3 & 18.3 & 96.3 \\
EOVSAM (Ours)                 & COCO Panoptic & \textbf{39.0} & \textbf{16.6} & 60.9 & \textbf{20.4} & \textbf{97.0} \\
\bottomrule
\end{tabular}
\caption{Open-vocabulary semantic segmentation performance. Models are grouped by their training datasets. Best results within each training dataset group are highlighted in bold.
$\dagger$ indicates our reproduced results.
}
\label{tab:OVSS performance}
\end{table*}

\begin{figure}[!t]
    \centering
    \includegraphics[width=1.0\linewidth]{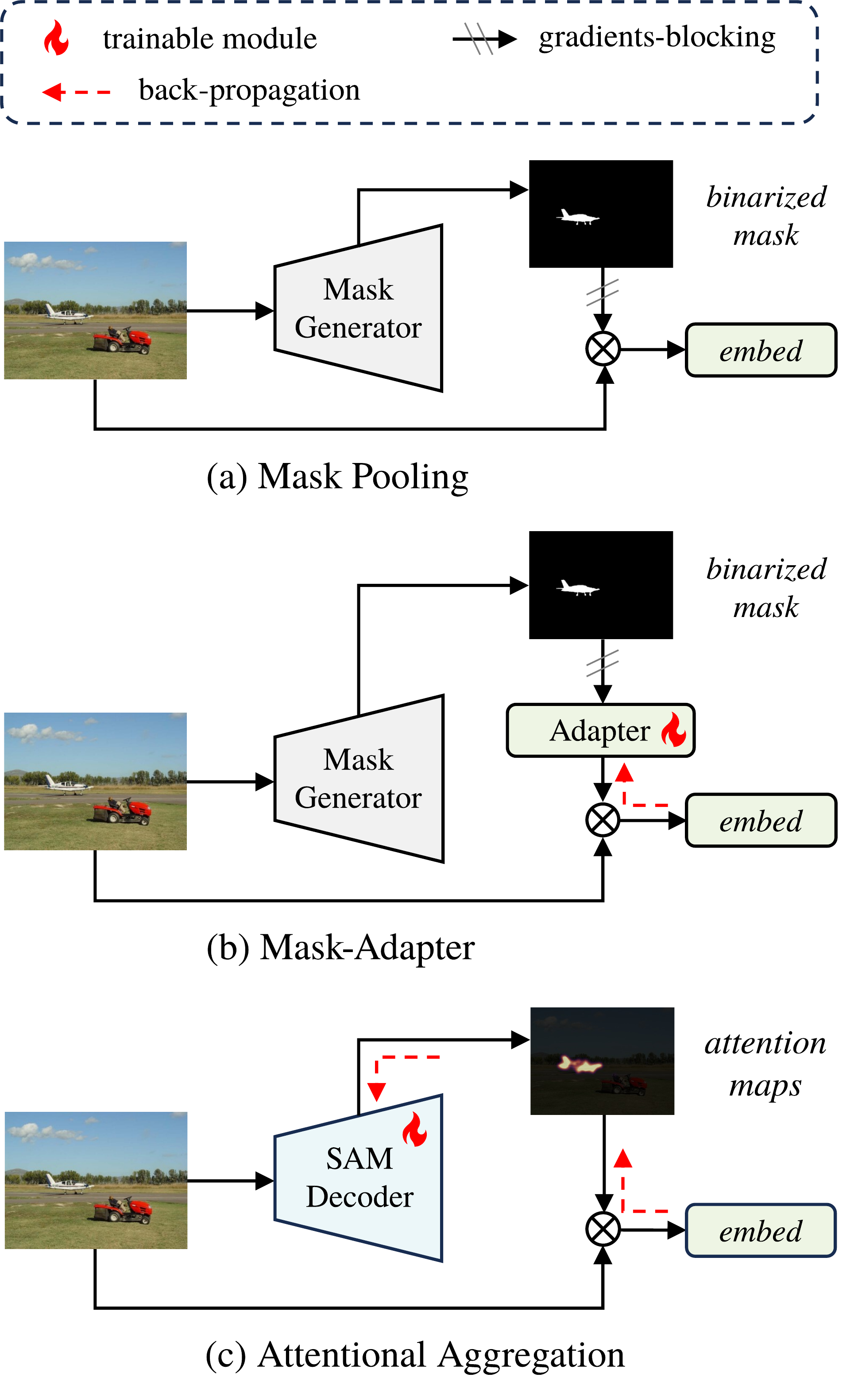}
    \caption{Comparison of feature aggregation strategies regarding gradient flow and computational overhead.
% (a) Mask Pooling: The non-differentiable nature of binarized masks blocks back-propagation, rendering the process completely non-trainable.
% (b) Mask-Adapter: Introduces extra adapter modules to enable training, but processing masks independently through these modules incurs massive computational and memory overhead.
% (c) Attentional Aggregation (Ours): Utilizes soft attention maps to maintain a continuous gradient flow for fully end-to-end joint optimization.
}
    \label{fig:pool_show}
\end{figure}

\subsection{Efficient Open-Vocabulary Recognition via Attentional Aggregation}

While enabling single-pass segmentation, we find that the SAM classifier generalizes poorly to unseen classes ($\mathcal{C}_\text{unseen}$) as it is optimized on a finite base vocabulary. To efficiently extract robust representations for novel concepts, we introduce an attentional feature aggregation strategy within the detector decoder.

The detector decoder consists of $L=6$ stacked transformer layers that process $N=200$ learnable object queries $\mathcal{Q}$, using the enhanced features $\mathcal{F}_\text{enh}$ as keys and values. To construct a classification-oriented attention map $\mathcal{A} \in \mathbb{R}^{N \times h \times w}$, we introduce a set of learnable weight coefficients for each attention head to pool the multi-head attention maps $\hat{\mathcal{A}}_l \in \mathbb{R}^{n_\text{head} \times N \times h \times w}$. Since the object queries $\mathcal{Q} \in \mathbb{R}^{N \times C}$ consistently interact with the enhanced features $\mathcal{F}_\text{enh}$ across all layers, the resulting $\mathcal{A}$ inherently maintains a precise point-to-point spatial correspondence with the visual features $\mathcal{F}_\text{siglip} \in \mathbb{R}^{C \times h \times w}$. Finally, we leverage the soft attention map $\mathcal{A}$ to dynamically aggregate $\mathcal{F}_\text{siglip}$, obtaining region-specific object embeddings $\mathcal{O}\in\mathbb{R}^{N\times C}$:
\begin{equation}
\mathbf{o}_i = \sum_{u,v} \mathcal{A}(i,u,v)  \mathcal{F}_{\text{siglip}}(:,u,v)
\label{eq:attentional_aggregation}
\end{equation}
where $(u, v)$ iterates over the spatial dimensions $h \times w$.
The main classification probability $\mathcal{P}_{i,k}$ for the $i$-th object belonging to the $k$-th class is derived from the cosine similarities between the object embedding $\mathbf{o}_i \in \mathcal{O}$ and the text embeddings $\mathbf{t}_k \in \mathcal{T}_\text{siglip}$:
\begin{equation}
\mathcal{P}_{i,k} = \frac{\exp(\cos(\mathbf{o}_i,\mathbf{t}_k)/\tau)}{\sum_{j=1}^{K}\exp(\cos(\mathbf{o}_i,\mathbf{t}_j)/\tau)}
\label{eq:classification_prob}
\end{equation}
where $\tau$ is a learnable temperature parameter and $\cos(\cdot, \cdot)$ denotes the cosine similarity function.

\begin{figure*}[t]
    \centering
    \includegraphics[width=1\linewidth]{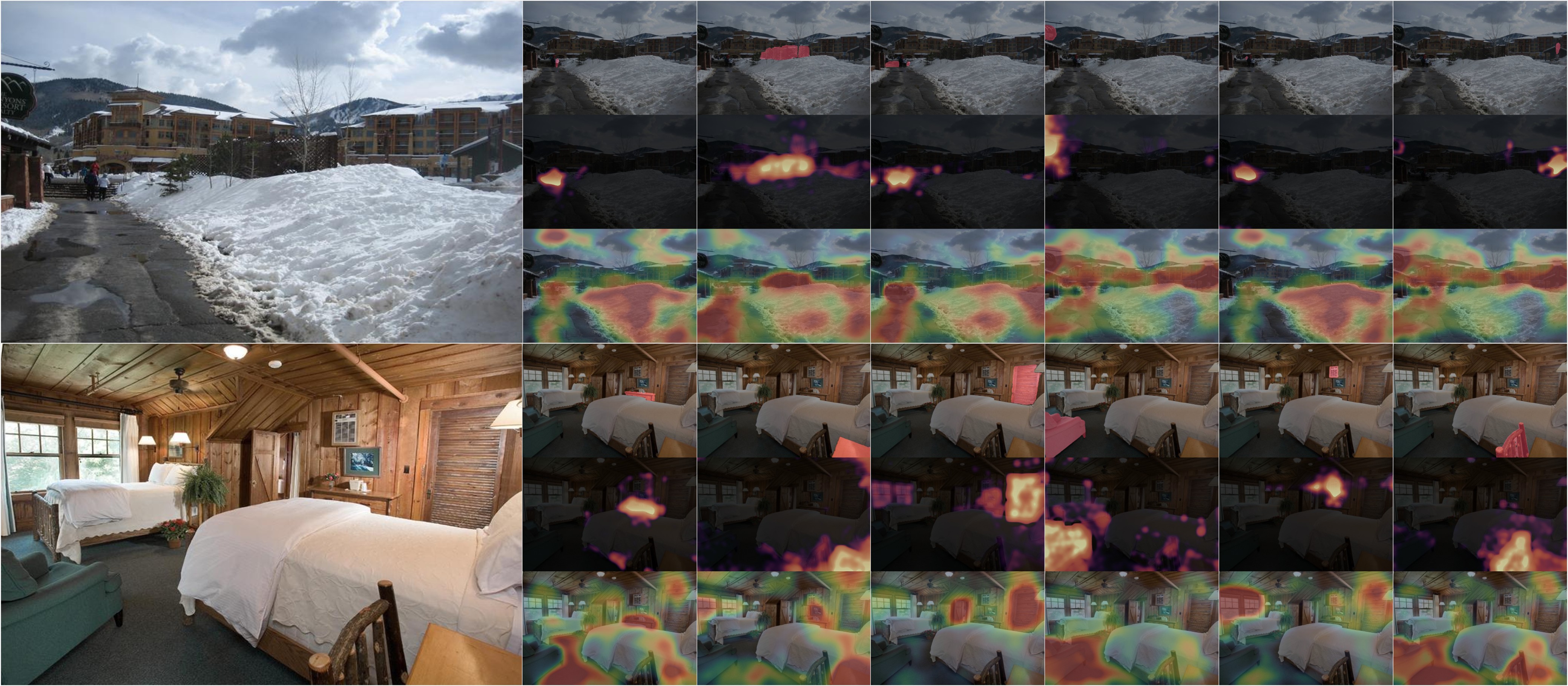}
    \caption{Qualitative results of our proposed model. We present visualizations of the masks and attention maps predicted by our model. For each image, the top row displays the predicted mask, while the second row exhibits the corresponding attention map. To further elucidate the function of the attention map, the bottom row illustrates the cosine similarity between the image features and the text features of the mask's corresponding category. 
    % We observe that the model successfully segments various objects within the image and focuses its attention maps on discriminative regions.
    }
    \label{fig:qualitative}
\end{figure*}

As shown in Figure~\ref{fig:pool_show}, our strategy fundamentally differs from prior mask-based extraction methods like mask cropping~\cite{simplebaseline,zegformer}, mask pooling~\cite{fcclip,maftplus}, or Mask-Adapter~\cite{maskadapter}. These previous methods predominantly rely on non-differentiable binarized masks, thereby truncating the gradient flow and hindering model optimization. Furthermore, instance-level sequential processing incurs prohibitive memory and computational overhead. By eliminating both sequential bottlenecks and non-differentiable operations, our method enables efficient, end-to-end joint optimization of mask generation and semantic representations within a single forward pass. Qualitative visualizations of our model's predictions are provided in Figure~\ref{fig:qualitative}.

\subsection{Objective}

Our optimization objective is built upon the loss functions established by Mask2Former~\cite{mask2former}. Specifically, this base loss consists of mask prediction losses (binary cross-entropy $\mathcal{L}_\text{bce}$ and Dice loss $\mathcal{L}_\text{dice}$) along with a classification loss, denoted here as $\mathcal{L}_\text{ce}$. We explicitly distinguish $\mathcal{L}_\text{ce}$ from our classification loss $\mathcal{L}_\text{cls}$: while $\mathcal{L}_\text{ce}$ primarily supervises object existence and exhibits poor generalization on unseen categories, $\mathcal{L}_\text{cls}$ is fully dedicated to open-vocabulary recognition. To compute $\mathcal{L}_\text{cls}$, we employ a cross-entropy loss based on the cosine similarities between the region-specific object embeddings $\mathcal{O}$ and the text embeddings $\mathcal{T}_\text{siglip}$. Additionally, to supervise the localization branch, we introduce $L_1$ loss ($\mathcal{L}_\text{L1}$) and generalized IoU loss ($\mathcal{L}_\text{giou}$) for bounding box prediction. Expanding all components, the total loss $\mathcal{L}_\text{total}$ is formulated as:
\begin{equation}
\begin{split}
\mathcal{L}_\text{total} &= \lambda_\text{ce}\mathcal{L}_\text{ce} + \lambda_\text{bce}\mathcal{L}_\text{bce} + \lambda_\text{dice}\mathcal{L}_\text{dice} \\
&\quad + \lambda_\text{cls}\mathcal{L}_\text{cls} + \lambda_\text{L1}\mathcal{L}_\text{L1} + \lambda_\text{giou}\mathcal{L}_\text{giou}
\end{split}
\label{eq:total_loss}
\end{equation}
We empirically set the weighting coefficients to $\lambda_\text{ce}=3$, $\lambda_\text{bce}=5$, $\lambda_\text{dice}=5$, $\lambda_\text{cls}=1$, $\lambda_\text{L1}=5$, and $\lambda_\text{giou}=2$.

\begin{table}
\centering
\small
\setlength{\tabcolsep}{3pt}
\begin{tabular}{lccc} % 1. 去掉竖线 (l|ccc 改为 lccc)
\toprule % 2. 顶部线（粗线）
\textbf{Method} & \textbf{PQ} & \textbf{SQ} & \textbf{RQ} \\ \hline
EOV-Seg~\cite{eovseg}         & 24.5        & 70.2        & 30.1        \\
FrozenSeg~\cite{frozenseg}      & 25.9        & -           & -           \\
FC-CLIP~\cite{fcclip}          & 26.8        & 71.5        & 32.3        \\ 
FC-CLIP w/ MaskAdapter$^\dagger$~\cite{maskadapter} & 26.6      & 71.3          & 32.1          \\
MAFT+~\cite{maftplus}           & 27.1        & 73.5        & 32.9        \\
% MROVSeg~\cite{mrovseg}         & 27.3        & 72.8        & 33.4        \\
OMTSeg~\cite{omtseg}          & 27.5        & -           & -           \\
% FISA~\cite{fisa}            & 28.1        & -           & -           \\
% MaskCLIP++/MAFT+~\cite{zeng2025maskclippp}& 28.1        & 74.0        & 34.7        \\
GBA~\cite{gba}             & 29.6        & 74.0        & 35.8        \\
OVRCOAT~\cite{ovrcoat}     & 28.6        & \textbf{77.3}    & 34.7        \\
\hline
SAM 3$^\dagger$~\cite{SAM_3} & 7.9       & 64.0             & 10.6      \\
EOVSAM (Ours)            & \textbf{30.9}        & 73.8        & \textbf{37.3}       \\ \hline
\end{tabular}
\caption{Open-vocabulary panoptic segmentation performance. We evaluate our method on the ADE20K dataset.
$\dagger$ indicates our reproduced results.
}
\label{tab:OVPS performance}
\end{table}

\section{Experiments}

\begin{table*}
    \centering
    \small
    \setlength{\tabcolsep}{2.0pt}
    \begin{tabular}{l | c c c | c c | c c | c c | c c}
        \toprule
        \multirow{2}{*}{\textbf{Method}}
        & \multicolumn{3}{c|}{\textbf{A-150}}
        & \multicolumn{2}{c|}{\textbf{A-847}}
        & \multicolumn{2}{c|}{\textbf{PC-59}}
        & \multicolumn{2}{c|}{\textbf{PC-459}}
        & \multicolumn{2}{c}{\textbf{PAS-20}} \\
        \cmidrule(lr){2-4}\cmidrule(lr){5-6}\cmidrule(lr){7-8}\cmidrule(lr){9-10}\cmidrule(lr){11-12}
        & mIoU & PQ & FPS & mIoU & FPS & mIoU & FPS & mIoU & FPS & mIoU & FPS \\
        \midrule
        FC-CLIP~\cite{fcclip}
        & 34.1 & 26.8 & 3.92 & 14.8 & 2.40 & 58.4 & 4.82 & 18.2 & 2.80 & 95.4 & 5.82 \\
        FC-CLIP w/ MaskAdapter~\cite{maskadapter}
        & 36.6 & 26.6 & 0.92 & 14.1 & 0.83 & 59.7 & 0.92 & 19.3 & 0.95 & 95.5 & 1.01 \\
        SAM 3$^\dagger$~\cite{SAM_3}
        & 37.8 & 7.9 & 0.03 & 14.0 & 0.01 & 58.3 & 0.07 & 18.3 & 0.02 & 96.3 & 0.14 \\
        \midrule
        EOVSAM (512)
        & 37.0 & 27.0 & \textbf{8.76} & 15.4 & \textbf{6.59} & 59.1 & \textbf{10.44} & 18.8 & \textbf{7.86} & 96.1 & \textbf{11.25} \\
        EOVSAM (576)
        & 37.9 & 28.9 & 8.66 & 15.8 & 5.95 & 59.6 & 9.12 & 19.3 & 6.34 & 96.4 & 9.21 \\
        % EOVSAM (640)
        % & 38.2 & 29.7 & 7.47 & 16.1 & 5.88 & 59.4 & 7.60 & 19.8 & 5.60 & 96.6 & 8.85 \\
        EOVSAM (768)
        & 38.4 & 30.0 & 6.06 & 16.2 & 5.32 & 59.9 & 6.85 & 19.7 & 5.42 & 96.8 & 6.71 \\
        % EOVSAM (896)
        % & 38.4 & 29.8 & 5.41 & 16.1 & 4.60 & 60.3 & 5.63 & 19.7 & 4.71 & 97.0 & 6.03 \\
        % EOVSAM (1024)
        % & 38.5 & 30.5 & 4.50 & 16.5 & 4.02 & 60.6 & 4.81 & 20.0 & 4.13 & 97.0 & 4.93 \\
        EOVSAM (1152)
        & 39.0 & 30.9 & 3.77 & 16.6 & 3.38 & 60.9 & 4.06 & 20.4 & 3.57 & 97.0 & 4.15 \\
        \bottomrule
    \end{tabular}
    \caption{Inference efficiency. The FPS of the compared methods is measured at their official resolutions. EOVSAM uses a single checkpoint across all resolutions, the numbers in parentheses indicate the input resolution scale. Results are measured on a single RTX 3090 GPU (CUDA 12.4), using identical class lists and precomputed text features.}
    \label{tab:efficiency_scaling}
\end{table*}

\subsection{Experimental Setup}
\paragraph{Datasets} 
We evaluate our method on widely used open-vocabulary segmentation benchmarks: COCO Panoptic~\cite{cocopano}, ADE20K~\cite{ade20k}, Pascal-Context~\cite{pascal_context}, and Pascal-VOC~\cite{Pascal-VOC}. For both semantic and panoptic segmentation~\cite{panoptic} tasks, we exclusively train EOVSAM on the COCO Panoptic dataset. To assess open-vocabulary semantic segmentation, the model is tested on ADE20K (A-847, A-150), Pascal-Context (PC-459, PC-59), and Pascal-VOC (PAS-20). For the open-vocabulary panoptic setting, the model is evaluated on ADE20K.

\paragraph{Evaluation Metrics} 
In order to quantitatively assess the model's performance, we adopt standard evaluation practices~\cite{fcclip,maftplus}. Semantic segmentation performance is evaluated using the mean Intersection over Union (mIoU). Additionally, panoptic segmentation is evaluated in terms of panoptic quality (PQ), segmentation quality (SQ), and recognition quality (RQ).

\paragraph{Implementation details}
We use NVIDIA's C-RADIOv4-H as the visual backbone, which is distilled from a set of teachers: SigLIP 2, DINOv3, and SAM 3. During training, the visual backbone, alongside the SAM 3 and SigLIP 2 text encoders, is completely frozen. The training is optimized with the AdamW~\cite{adamw} optimizer and a weight decay of $1 \times 10^{-4}$. We use a crop size of 1152 $\times$ 1152. We employ an initial learning rate of $1 \times 10^{-4}$ and utilize a multi-step learning rate decay schedule. This is preceded by a linear warmup phase for the first 2,500 iterations, starting with a warmup factor of 0.001. The model is trained for 20,000 iterations on the COCO Panoptic training set with a batch size of 8. All training experiments are conducted on 2 NVIDIA RTX 3090 GPUs. During inference, the longer side of input images is resized to 1152.

\subsection{Results}
In this section, we evaluate our proposed EOVSAM using vanilla SAM 3 as the primary baseline. We further compare our approach with state-of-the-art open-vocabulary semantic and panoptic segmentation methods. All results are obtained from a single model trained on COCO Panoptic.

\paragraph{Open-Vocabulary Semantic Segmentation}
We follow standard evaluation protocols~\cite{fcclip} for open-vocabulary semantic segmentation. The COCO-Stuff~\cite{cocostuff} dataset comprises 118k images across 171 categories, whereas COCO Panoptic shares the same training images but encompasses 133 categories. Given that models trained on COCO-Stuff generally outperform those trained on COCO Panoptic in open-vocabulary semantic segmentation tasks~\cite{maftplus,maskadapter} but inherently lack the capability for panoptic segmentation, we categorize the evaluated models based on their training datasets.
Despite being trained exclusively on COCO Panoptic, our model exhibits strong open-vocabulary semantic segmentation capabilities. Table~\ref{tab:OVSS performance} details the performance of EOVSAM across various benchmarks. Compared to SAM 3, EOVSAM demonstrates significant improvements, yielding mIoU gains of +1.2, +2.6, +2.6, +2.1 and +0.7 on A-150, A-847, PC-59, PC-459, and PAS-20, respectively. Notably, among all models trained on COCO Panoptic, our approach achieves state-of-the-art results on A-150, A-847, PC-459 and PAS-20. Furthermore, EOVSAM delivers performance comparable to leading open-vocabulary semantic segmentation models trained on COCO-Stuff.

\paragraph{Open-Vocabulary Panoptic Segmentation}
As detailed in Table~\ref{tab:OVPS performance}, we evaluate the proposed EOVSAM on ADE20K, the primary benchmark dataset for open-vocabulary panoptic segmentation. Notably, EOVSAM significantly outperforms previous works on both the PQ and RQ metrics. 
% This demonstrates its robust comprehensive segmentation capabilities and validates the effectiveness of the Attentional Aggregation.

\paragraph{Inference Efficiency.}
Given that FC-CLIP~\cite{fcclip}, MAFT+~\cite{maftplus}, and OVRCOAT~\cite{ovrcoat} share similar architectural designs and inference speeds, we select FC-CLIP to represent the speed characteristics of these approaches.
As shown in Table~\ref{tab:efficiency_scaling}, EOVSAM maintains robust performance at reduced resolutions and achieves further acceleration without retraining. 
% This flexibility is particularly beneficial for latency-sensitive and resource-constrained deployment.
Notably, our EOVSAM outperforms SAM 3 even at a resolution of 576.

% 放补充

% \begin{table}[t]
%   \centering
%   \small
%   \begin{tabular}{lcc}
%     \toprule
%     \textbf{Method} &  \textbf{A-150} & \textbf{A-847} \\
%     \midrule
%     FC-CLIP~\cite{fcclip} & 2.70 & 2.65 \\
%     FC-CLIP w/ MaskAdapter~\cite{maskadapter} & 1.59 & 1.55 \\
%     SAM 3~\cite{SAM_3} & 0.03 & 0.01 \\
%     EOVSAM (Ours) & \textbf{3.77} & \textbf{3.38} \\
%     \bottomrule
%   \end{tabular}
%   \caption{FPS comparison. All results are evaluated on a single NVIDIA RTX 3090 GPU with CUDA 12.4. For all models, we use the same list of class names.}
%   \label{tab:FPS}
% \end{table}

% \paragraph{Inference Speed.}
% In Table~\ref{tab:FPS}, we compare the FPS (frames per second) of EOVSAM with SAM 3 and other state-of-the-art methods. Our approach achieves significantly faster inference speeds and does not suffer from substantial speed degradation when the number of target categories increases.

\subsection{Additional Experiments}

\begin{table}[t]
    \centering
    \small
    \setlength{\tabcolsep}{2.5pt}
    \begin{tabular}{l | c c c c c c}
        \toprule
        \multirow{2}{*}{\textbf{Variants}} & \multicolumn{2}{c}{\textbf{A-150}} & \textbf{A-847} & \textbf{PC-59} & \textbf{PC-459} & \textbf{PAS-20} \\
        \cmidrule(lr){2-3} 
        & mIoU & PQ & mIoU & mIoU & mIoU & mIoU \\
        \midrule
        w/o SAM 3 Weights & 35.2 & 24.2 & 15.2 & 58.9 & 18.7 & 95.5 \\
        w/o Attn. Agg.    & 22.6 & 19.8 & 5.1  & 59.6 & 12.4 & 94.9 \\
        w/o Box Sup.      & 38.2 & 30.4 & 16.0 & \textbf{61.2} & 19.7 & 96.5 \\
        w/ Text C-A       & 37.9 & 29.9 & 16.1 & 60.8 & 19.6 & 96.4 \\
        \midrule
        EOVSAM (Full)     & \textbf{39.0} & \textbf{30.9} & \textbf{16.6} & 60.9 & \textbf{20.4} & \textbf{97.0} \\
        \bottomrule
    \end{tabular}
    \caption{Ablation study on components of EOVSAM. ``SAM 3 Weights'' indicates loading the pre-trained SAM 3 weights. ``Attn. Agg.'' denotes Attentional Aggregation, ``Box Sup.'' refers to Bounding Box Supervision, and ``Text C-A'' stands for Text Cross-Attention.}
    \label{tab:ablation}
\end{table}

\paragraph{Component-Level Ablation Studies}
Table~\ref{tab:ablation} validates the key components of EOVSAM. First, discarding the pre-trained SAM 3 weights causes a drastic performance drop ($-6.7$ PQ and $-3.8$ mIoU on A-150), demonstrating that inheriting SAM 3's localization priors is essential for precise segmentation. Second, while the variant without Attentional Aggregation performs acceptably on datasets similar to the training domain (such as PC-59 and PAS-20), it collapses on complex, out-of-distribution scenes. Our Attentional Aggregation effectively bridges this gap, yielding massive generalization gains (up to +16.4 mIoU and +11.1 PQ). Third, auxiliary bounding box supervision provides beneficial explicit spatial guidance across most domains. Finally, removing text cross-attention to create a prompt-free architecture not only eliminates the computational bottleneck but also marginally improves overall accuracy.

\begin{table}
    \centering
    \small
    \setlength{\tabcolsep}{2.0pt}
    \begin{tabular}{l | c c c c c c c}
        \toprule
        % 第一层表头：在第8列添加 \multirow{2}{*}{FPS} 以跨越两行
        & \multicolumn{3}{c}{\textbf{A-150}} & \textbf{A-847} & \textbf{PC-59} & \textbf{PC-459} & \textbf{PAS-20}  \\
        
        \cmidrule(lr){2-4} 
        
        & mIoU & PQ & FPS & mIoU & mIoU & mIoU & mIoU  \\
        \midrule
        Mask Pooling      & 36.0 & 28.9 & \textbf{3.80} & 15.7 & 59.7 & 18.8 & 96.3 \\
        \midrule
        Mask-Adapter      & 37.3 & 29.8 & 1.68 & 15.8 & 60.3 & 18.8 & 96.3 \\
        \midrule
        Attn. Agg. & \textbf{39.0} & \textbf{30.9} & 3.77 & \textbf{16.6} & \textbf{60.9} & \textbf{20.4} & \textbf{97.0} \\
        \bottomrule
    \end{tabular}
    \caption{Comparison of Attentional Aggregation with Other Methods. For Mask Pooling and Mask-Adapter, we remove Attentional Aggregation from EOVSAM and utilize the remaining components as a mask generator.}
    \label{tab:aggregate}
\end{table}

\paragraph{Comparison of Attentional Aggregation with Other Methods}
Table~\ref{tab:aggregate} compares various feature aggregation methods. Our approach outperforms the others in terms of accuracy while maintaining high computational efficiency comparable to that of mask pooling.

\begin{table}[t]
    \centering
    \small
    \setlength{\tabcolsep}{2.5pt}
    % 现在的列数变成了：2列参数 + 6列数据 = 8列
    % c c 代表 alpha 和 beta 居中对齐，| 依然作为参数和指标的分割线
    \begin{tabular}{c c | c c c c c c}
        \toprule
        % 第一层表头：使用 multirow 让 \alpha 和 \beta 跨两行并垂直居中
        % 第一层表头：使用 bm 加粗希腊字母
        \multirow{2}{*}{$\bm{\alpha}$} & \multirow{2}{*}{$\bm{\beta}$} & \multicolumn{2}{c}{\textbf{A-150}} & \textbf{A-847} & \textbf{PC-59} & \textbf{PC-459} & \textbf{PAS-20} \\
        
        \cmidrule(lr){3-4} 
        
        % 第二层表头：前两列留空（因为已经被 multirow 占据），后面跟具体指标
        & & mIoU & PQ & mIoU & mIoU & mIoU & mIoU \\
        \midrule
                    0.0 & 0.0 & 22.3 & 20.2 & 5.1 & 59.5 & 13.1 & 95.8 \\
                    0.0 & 1.0 & 35.5 & 27.8 & 15.2 & 60.6 & 18.8 & 96.0 \\
                    0.4 & 0.8 & 35.1 & 27.1 & 14.2 & 61.4 & 18.5 & 96.4 \\
                    % 0.5 & 0.5 & 30.8 & 24.3 & 10.5 & 61.5 & 17.5 & 96.5 \\
                    0.6 & 1.0 & 38.8 & 30.8 & 16.6 & 60.8 & 20.2 & 96.5 \\
                    0.7 & 1.0 & 39.0 & 30.9 & 16.6 & 60.9 & 20.4 & 97.0 \\
                    % 0.8 & 0.8 & 36.1 & 28.7 & 14.8 & 61.2 & 19.2 & 96.6 \\
                    0.8 & 1.0 & 38.8 & 31.0 & 16.7 & 60.9 & 20.5 & 97.0 \\
                    % 0.9 & 1.0 & 38.8 & 31.1 & 16.8 & 60.9 & 20.4 & 97.2 \\
                    1.0 & 1.0 & 38.4 & 30.9 & 16.7 & 60.6 & 20.3 & 97.2 \\
        \bottomrule
    \end{tabular}
    \caption{Impact of geometric ensembling coefficients. $\alpha$ and $\beta$ denote the exponential weights assigned to the SigLIP classifier for seen and unseen categories, respectively, during the geometric ensembling with the SAM classifier.}
    \label{tab:ablation_coeff}
\end{table}

\paragraph{Coefficient Selection for Geometric Ensembling} Following~\cite{fcclip}, we employ a geometric ensemble strategy on the outputs of the SAM and SigLIP classifiers to determine the final categories. As shown in Table~\ref{tab:ablation_coeff},
% compared to the results reported in~\cite{fcclip},
our method achieves strong performance even without this ensemble (using only the SigLIP classifier, $\alpha=\beta=1.0$), demonstrating strong generalization. For simplicity, we adopt a unified setting ($\alpha=0.7, \beta=1.0$) across all datasets, proving robustness without dataset-specific tuning.

% \begin{table}
%     \centering
%     % l 代表左对齐，c 代表居中对齐，| 代表竖线
%     \small
%     \setlength{\tabcolsep}{2.6pt}
%     \begin{tabular}{l | c c c c c}
%         \toprule
%         \textbf{Training dataset} & \textbf{A-150} & \textbf{A-847} & \textbf{PC-59} & \textbf{PC-459} & \textbf{PAS-20} \\
%         \midrule
%         Stuff    & 37.5 & 13.2 & \textbf{62.0} & 18.3 & \textbf{97.6} \\
%         \midrule
%         Panoptic (w/o box sup.) & 38.2 & 16.0 & 61.2 & 19.7 & 96.5\\
%         \midrule
%         Panoptic & \textbf{39.0} & \textbf{16.6} & 60.9 & \textbf{20.4} & 97.0 \\ 
%         \bottomrule
%     \end{tabular}
%     \caption{Impact of training datasets. Comparison between models trained on COCO-Stuff and COCO Panoptic.}
%     \label{tab:performance_dataset}
% \end{table}

% \paragraph{Impact of Annotation Granularity}
% As shown in Table~\ref{tab:performance_dataset}, COCO-Stuff yields sub-optimal performance compared to COCO Panoptic. To isolate the impact of bounding boxes absent in COCO-Stuff, we evaluate a box-free panoptic variant. Crucially, this variant still maintains an overall advantage. We hypothesize that as an instance-level foundation model, SAM 3 benefits more from the instance-centric supervision of COCO Panoptic than the category-level annotations of COCO-Stuff.

\section{Conclusion}
In this work, we presented EOVSAM, an efficient adaptation of SAM 3 for single-pass open-vocabulary segmentation.
By converting SAM 3 into a prompt-free mask generator, EOVSAM inherits its strong localization priors while eliminating the multi-pass computational bottleneck. Additionally, it introduces a novel Attentional Aggregation strategy to optimize open-vocabulary classification end-to-end. EOVSAM improves both accuracy and inference speed over vanilla SAM 3, achieving speedups of up to 338$\times$.
Moreover, EOVSAM delivers highly competitive accuracy at reduced input resolutions without the need for resolution-specific fine-tuning, affording a highly flexible trade-off between performance and latency.
Evaluations on standard benchmarks demonstrate that EOVSAM achieves superior speed and performance over SAM 3 and other state-of-the-art methods, establishing a strong baseline for future research.

\bibliography{eovsam}

\end{document}